\documentclass{article}

\usepackage{microtype}
\usepackage{graphicx}
\usepackage{subfigure}
\usepackage{xcolor}
\usepackage{booktabs}
\usepackage{hyperref}
\usepackage{amssymb}
\usepackage{amsmath}
\usepackage{amsthm}

\usepackage[accepted]{icml_arxiv}

\icmltitlerunning{Generating and Aligning from Data Geometries with Generative Adversarial Networks}

\begin{document}

\twocolumn[
\icmltitle{Generating and Aligning from Data Geometries with Generative Adversarial Networks}
\icmlsetsymbol{equal}{*}

\begin{icmlauthorlist}
\icmlauthor{Matthew Amodio}{yalecs}
\icmlauthor{Smita Krishnaswamy}{yalecs,yalegen}
\end{icmlauthorlist}

\icmlaffiliation{yalecs}{Department of Computer Science, Yale University}
\icmlaffiliation{yalegen}{Department of Genetics, Yale University}

\icmlcorrespondingauthor{Smita Krishnaswamy}{smita.krishnaswamy@yale.edu}
\icmlkeywords{deep learning, gan}

\vskip 0.3in
]
\printAffiliationsAndNotice{}  

\title{}
\author{Matthew Amodio}

\begin{abstract}
    Unsupervised domain mapping has attracted substantial attention in recent years due to the success of models based on the cycle-consistency assumption. These models map between two domains by fooling a probabilistic discriminator, thereby matching the probability distributions of the real and generated data. Instead of this probabilistic approach, we cast the problem in terms of aligning the geometry of the manifolds of the two domains. We introduce the Manifold Geometry Matching Generative Adversarial Network (MGM GAN), which adds two novel mechanisms to facilitate GANs sampling from the geometry of the manifold rather than the density and then aligning two manifold geometries: (1) an importance sampling technique that reweights points based on their density on the manifold, making the discriminator only able to discern geometry and (2) a penalty adapted from traditional manifold alignment literature that explicitly enforces the geometry to be preserved. The MGM GAN leverages the manifolds arising from a pre-trained autoencoder to bridge the gap between formal manifold alignment literature and existing GAN work, and demonstrate the advantages of modeling the manifold geometry over its density.
\end{abstract}

\section{Introduction}
Recently, generative adversarial networks (GANs) have emerged as a scalable solution for the generation of a wide variety of data types from images to text to biological samples~\cite{zhu2016generative,huang2017manifold}. GANs mainly aim to generate data distributed similarly as the training data, achieved with an alternating minimax game between a generator and a probabilistic discriminator. However, in many applications, simply replicating the density is not very useful~\cite{grun2015single}. This is because the process of collecting data is often biased, and there is uneven coverage of the data state space, leaving important areas of the state space sparsely sampled~\cite{lindstrom2011miniaturization}. This problem is exacerbated when aligning datasets from different batches, where the goal is to match the geometries of each dataset such that latent objects like cell types are aligned. GANs may skew the alignment of the datasets based on differences in density. To address this, we develop a geometry-based data generation and matching GAN framework.

Generating from a manifold geometry model of the data can be much more useful than replicating density in many applications~\cite{lindenbaum2018geometry,coifman2005geometric}. Such generation can i) ``fill'' missing samples from parts of the state space while ii) creating additional samples in sparse areas with rare data points that may be of great interest to the application area, such as biology. Modeling the manifold geometry traditionally involves converting the data into a k-nearest neighbors graph (or some other density-independent graph) and then computing the eigen dimensions of the associated graph Laplacian~\cite{belkin2008towards}. Other variations include turning the graph Laplacian into a random walk operator and diffusing over the graph~\cite{belkin2008towards}. In either case, this can become computationally infeasible in cases of large datasets due to the complexity of eigendecomposition and graph computation. Furthermore, it is often not known how to compute meaningful distances when Euclidean distance is inappropriate. For example, a simple pixel-by-pixel L2 distance between the two images is not an accurate measure of how similar they are, and developing meaningful distances measures remain an ongoing challenge~\cite{frogner2015learning,wang2005euclidean}.

\begin{figure*}
    \centering
    \includegraphics[width=\linewidth]{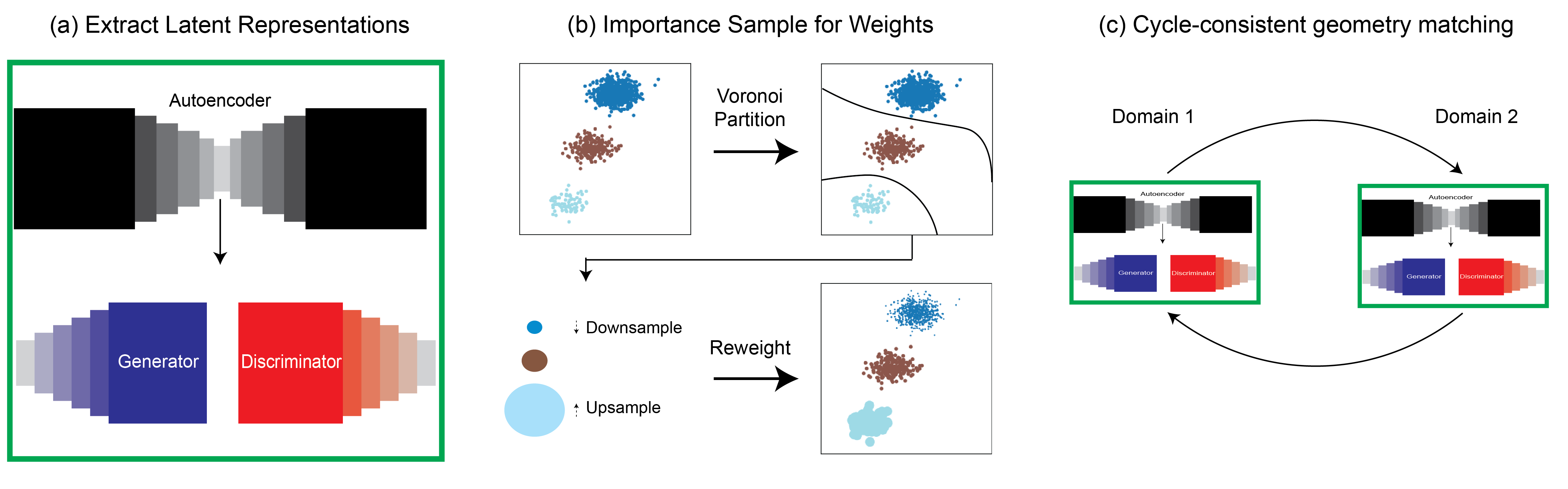}
    \caption{(a) The Manifold Geometry (MG) GAN architecture first extracts the latent representations of the data from an autoencoder to approximate the data manifold, (b) and then reweights points with an importance sampling scheme that balances the densities along the manifold. (c) Two MG GANs in a cycle-consistent framework form the Manifold Geometry Matching (MGM) architecture for unsupervised domain mapping.}
    \label{fig:cartoon1}
\end{figure*}

Previous deep learning literature has shown that autoencoders are capable of learning the manifold geometry of a dataset without placing any assumptions on distance~\cite{vincent2010stacked,vincent2008extracting,amodio2017exploring,holden2015learning}. Here we propose to augment the two network system of a GAN (generator and discriminator) with a third network (an autoencoder) to generate data from the manifold geometry. We call this new framework the Manifold Geometry GAN (MG GAN), which has a new manifold geometry loss function. We show that this loss is essential for aligning manifolds. Density-based alignment is very problematic and can skew the alignment between samples with density differences. We further show how to incorporate the geometric loss into a cycle-consistent GAN framework to form the Manifold Geometry Matching GAN (MGM GAN). The main contributions of our work are as follows:
\begin{enumerate}
    \item a novel loss that facilitates using a GAN to sample from the manifold geometry
    \item the cycle-consistent alignment framework of the MGM GAN
    \item demonstration of the difference between density generation and geometry generation on many datasets
\end{enumerate}

\section{Previous Work}
Much work has been devoted to the topic of modeling manifold geometry, largely focused on graph and distance based methods~\cite{tenenbaum2000global,lindenbaum2018geometry,wang2013manifold}. These include Laplacian eigenmaps and diffusion maps~\cite{tu2012laplacian,belkin2003laplacian,ellingson2010validation,coifman2005geometric}. The graph-based approach of \cite{oztireli2010spectral} balances density through a resampling scheme of existing data points. With a model of the manifold geometry, two manifolds can be aligned through other, graph-based harmonic methods~\cite{bachmann2005exploiting, stanley2018manifold}. Unlike this work, these methods all face the difficulties of forming a meaningful graph representation of the data discussed above.

Existing research has demonstrated autoencoders to be effective at learning the data manifold, both in theory and practice, without placing restrictive assumptions like a distance measure between points~\cite{vincent2010stacked,vincent2008extracting,holden2015learning,amodio2017exploring,bengio2013representation,goodfellow2016deep}. Augmented forms of autoencoders, including those with added adversaries, are a subject of continuing research in order to further leverage the powerful representations that autoencoders are able to learn~\cite{makhzani2015adversarial,wang2014generalized,wang2016auto}.

GANs have been used previously for domain mapping, with the earliest methods requiring supervision ~\cite{isola2017image,van2015transfer}. Unsupervised domain mapping came to the forefront with the introduction of cycle-consistent GANs. Cycle-consistent GANs address the problem of unsupervised domain mapping by simultaneously learning two generative functions: one that maps from the first dataset to the second dataset and vice versa. Many such pairs of generative functions exist, however, so the key assumption constraining these architectures is that the two generative functions should be each other's inverse. In practice, cycle-consistent GANs have achieved impressive results on a wide range of applications~\cite{hoffman2017cycada,zhu2017unpaired,chu2017cyclegan}. However, problems with the density-based loss for domain mapping, including model ambiguity, have been identified~\cite{amodio2018magan,dumoulin2016adversarially,perera2018in2i,li2017alice}.

Importance sampling is a frequently used technique in statistical literature, used for improving Monte Carlo simulations, Bayesian inference, and surveying. ~\cite{neal2001annealed,guo2002survey,gregoire2007sampling}. In stratified survey sampling, known biases in the sampling process are corrected by weighting populations differently~\cite{nassiuma2000survey}. It has only been rarely used in deep learning contexts, however, mostly with a focus on optimizing convergence~\cite{bengio2003quick}.

\begin{figure}
    \centering
    \includegraphics[width=\linewidth]{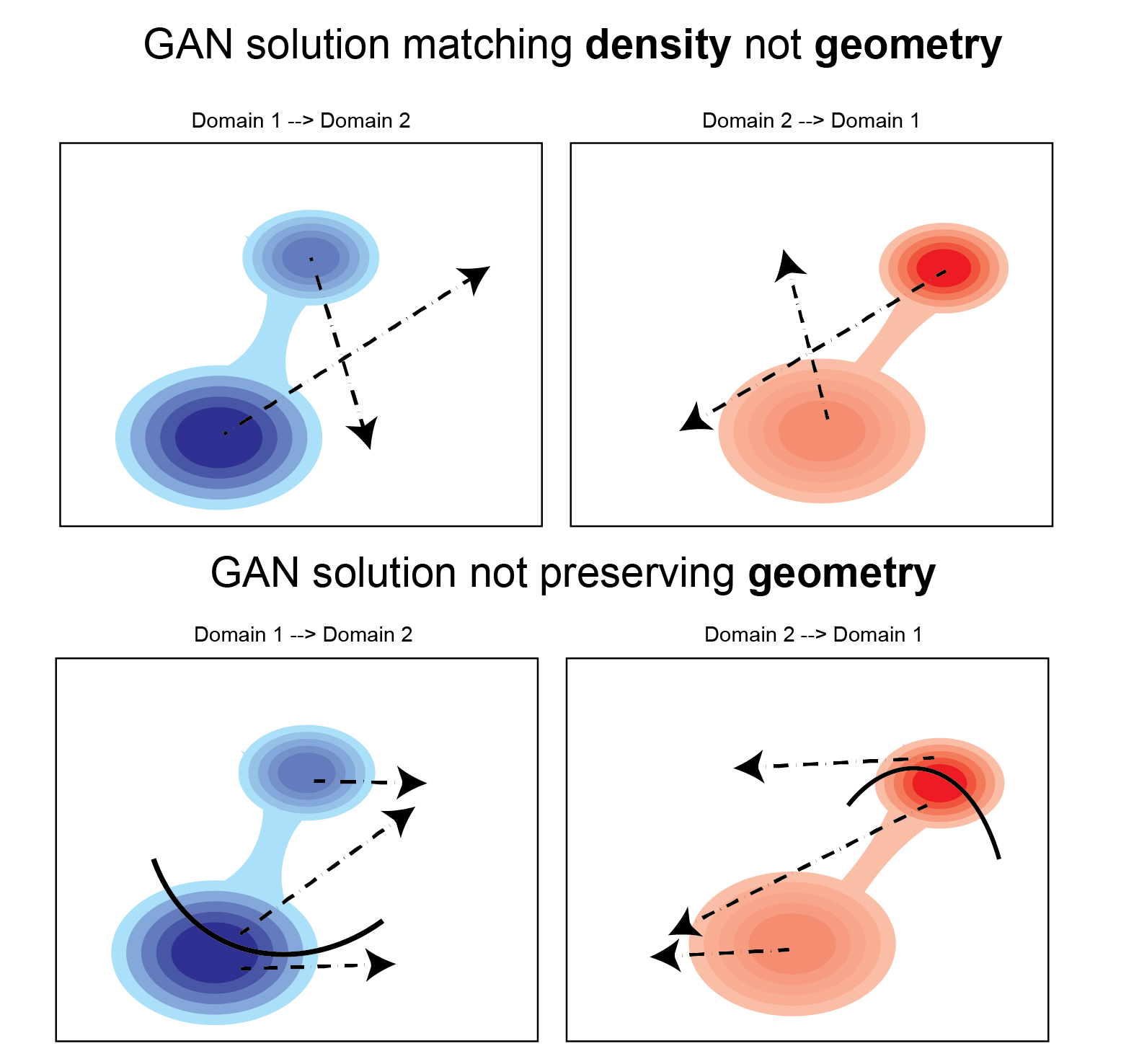}
    \caption{Two examples of a GAN not aligning data manifolds when underlying populations are sampled at different frequencies in the two domains. (a) The traditional GAN loss forces density to be matched rather than geometry. (b) The generator can sever the manifold geometry, mapping two similar points to two very different targets or two different points to two very similar targets.}
    \label{fig:cartoon2}
\end{figure}

\section{MGM GAN Model}
To model the geometry of a single manifold with a GAN we first let $X$ be a dataset domain with $x_i \in \mathbb{R}^{D_X}$, $i=1...N_X$. We seek a generator $G$ that takes points from a noise domain $z$ and maps them to $X$. To guide the generator into creating realistic points, we also train a discriminator network $D_X$ that tries to distinguish between real points and points mapped by the generator. Adversarial training leads the generator to try to fool the discriminator into classifying its points as real, with the following standard loss terms:
\begin{align}
    L_D &= \mathbb{E}[D_X ( G (z) ) ] - \mathbb{E}[D_X ( X ) ] \nonumber \\
    L_G &= - \mathbb{E}[D_X ( G (z) ) ] \nonumber
\end{align}

\subsection{Importance Sampling}
To make the GAN model the geometry instead of the density, we first obtain a representation of the data manifold $M$ by extracting a latent layer of a pre-trained autoencoder, letting $M_{x_i}$ be the representation of $x_i$ on the manifold (Figure~\ref{fig:cartoon1}a). We then create a Voronoi partition of $M$ with k-means clustering, dividing the space into $k$ regions $r_1...r_k$. We assign weights $w$ to each point inversely proportional to the number of points in its dataset that are in that the region on the manifold (Figure~\ref{fig:cartoon1}b):
\begin{align}
    w_i =& \sum_{j=1}^{k} \frac{\mathbb{I}_{x_i \in r_j}}{N_{X, r_j}} \nonumber \\
    N_{X, r_j} =& \sum_{i=1}^{N_X} \mathbb{I}_{x_i \in r_j} \nonumber
\end{align}



While these weights can be used to make a single generator sample from the geometry of a single domain manifold, we can also extend this to the case of unsupervised domain mapping where we have two datasets ($X$ and $Y$), and two generators ( $G_{XY}$ and $G_{YX}$) and discriminators ($D_X$ and $D_Y$), as well. Normally, the minimax game between the generator and the discriminator finds equilibrium when the discriminator's probabilistic output calculates each point as being equally likely to be from the real sample and the generated sample. In order to accomplish this, if a shared region of the space $r$ is sparser in $X$ than it is in $Y$, the generator $G_{XY}$ must take points in $X$ that are not in $r$ and map them to points in $r$. In other words, the generators are not learning to align the manifolds of $X$ and $Y$, but are learning to match the density in the data space.

This alignment warps the manifold geometry, taking points that were originally not similar (points in $r$ and points not in $r$), and projecting them close to each other. Since $r$ is already represented in the dataset, this region need not be altered at all. Instead, we want the generator to only change points if they do not look like realistic points in the other dataset. To accomplish this, we adopt the following importance sampling technique.

\textbf{Theorem} Under the traditional GAN loss, a generated distribution $G$ that occupies the same regions as the real distribution $X$ but with different densities is not a minimum.

\begin{proof} Without loss of generality, consider the discriminator loss for a small region \textit{r} such that $D_X (x_i) = D_X (x_j) \forall x_i, x_j \in r$:

\begin{align}
    L_{D} &= - \sum_{G_{YX}(y) \in r} D_X ( G_{YX} (y) )  - \sum_{x \in r} ( 1 - D_X ( x ) ) \nonumber \\
    L_{D} &= - \sum_{G_{YX}(y) \in r} D_X  - \sum_{x \in r} (1 - D_X) \nonumber
\end{align}
\begin{figure*}
    \centering
    \includegraphics[width=.7\linewidth]{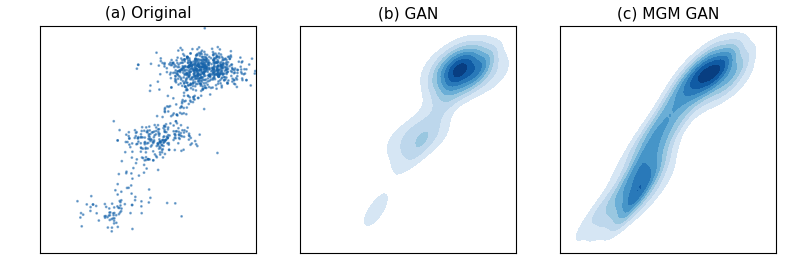}
    \caption{GANs can only to learn to generate from the density, which in many applications is less important than the geometry (for example, in biology where rare cells are just as, or even more, important for understanding the system). In this data, there are three populations of interest, and the MGM GAN is able to generate from the geometry, rather than the density, of the data manifold.}
    \label{fig:artificial2}
\end{figure*}

Using the notation from above where $N_{X, r}$ and $N_{G_{YX}(Y), r}$ are the number of points of $X$ and $G_{YX}(Y)$ that are in $r$, respectively:
\begin{align}
    L_D &= - D_X \cdot N_{G_{YX}(y), r} - (1 - D_X) \cdot N_{X, r} \nonumber \\
    L_D &= - D_X \cdot N_{G_{YX}(y), r} + D_X \cdot N_{X, r} - N_{X, r} \nonumber 
\end{align}

Next we take the derivative, and assuming that it is a local optimum, set it equal to zero:
\begin{align}
    0 &= - D_X^\prime \cdot N_{G_{YX}(y), r} + D_X^\prime \cdot N_{X, r} \nonumber \\
    D_X^\prime \cdot N_{G_{YX}(y), r} &= D_X^\prime \cdot N_{X, r} \nonumber \\
    N_{G_{YX}(y), r} &= N_{X, r} \nonumber
\end{align}

In general, the number of points in the real data $X$ and the generated data $G_{YX}(Y)$ that are in the region $r$ will not be the same, and thus by contradiction this is not a local optimum.
\end{proof}

\textbf{Theorem} Under the importance sampling GAN loss, a generated distribution $G$ that occupies the same regions as the real distribution $X$, even with different densities, is a minimum.

\begin{proof} As before, we will consider the discriminator loss for a small region $r$, except now with the importance sampling weights:

\begin{align}
    L_{D} &= - \sum_{G_{YX}(y) \in r} \frac{D_X ( G_{YX} (y) )}{N_{G_{YX}(y), r}}  - \sum_{x \in r} \frac{( 1 - D_X ( x ) )}{N_{X,r}} \nonumber \\
    L_{D} &= - \sum_{G_{YX}(y) \in r} \frac{D_X}{N_{G_{YX}(y), r}}  - \sum_{x \in r} \frac{( 1 - D_X )}{N_{X,r}} \nonumber \\
    L_D &= -  \frac{N_{G_{YX}(y), r} \cdot D_X}{N_{G_{YX}(y), r}} + \frac{N_{X,r} \cdot D_X}{N_{X,r}} - \frac{N_{X,r}}{N_{X,r}} \nonumber
\end{align}

\begin{figure*}
    \centering
    \includegraphics[width=\textwidth]{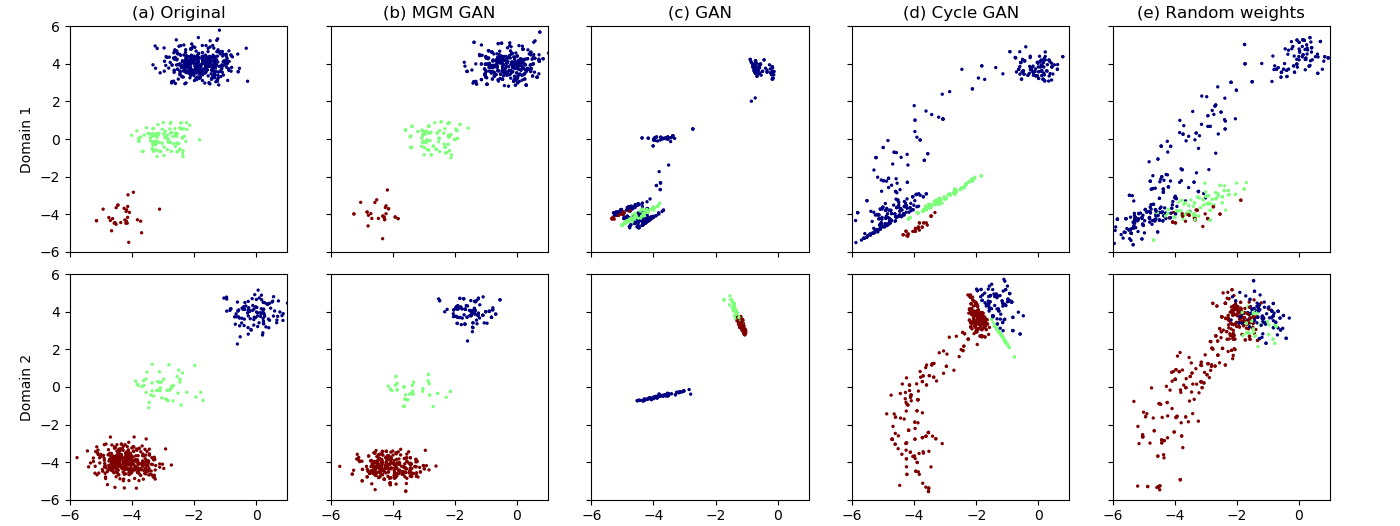}
    \caption{Simulated data consisting of a mixture of three Gaussians. Only one of the Gaussians was shifted between domains (dark blue), but differential sampling rates prevents traditional GANs from optimally aligning the two domain manifolds by their geometry. The MGM GAN's importance sampling makes it robust to these changes in density.}
    \label{fig:artificial}
\end{figure*}

For any values of $N_{X, r}$ and $N_{G_{YX}(y), r}$ greater than zero, this function is a constant and thus its derivative is zero. Thus, the importance sampling GAN loss in a region has a local optimum anywhere there are both real and generated points in that region, no matter their respective densities.
\end{proof}

Note that the modified loss function cannot be further lowered for any region that has points in both the real and generated data, but since the weights are constant with respect to the network, the function is still fully differentiable and receives signal to generate points in regions where there is real data but where it does not currently output any points. We further note that while this method requires choosing the number of partitions $r$, we find that optimizing the Bayesian Information Criterion (BIC) over partitions of the manifold works well~\cite{Konishi2008}.

Under the importance sampling loss function, we have altered the loss landscape such that mappings that match the geometry of the two manifolds are optima. Now that these mappings are local optima, the GAN framework \textit{could potentially} succeed at matching manifold geometry. In the traditional framework, optimization would be guaranteed \textit{not} to find these mappings, since they were not local optima. However, this loss alone does not explicitly enforce manifold alignment. Thus, we also introduce a global manifold alignment loss term for this purpose.

\subsection{Manifold Geometry Loss}
Preserving the manifold geometry requires preserving some notion of distance between points before and after transformation. However, the standard GAN loss function only looks at the data after transformation, and thus cannot enforce any relationship between points before and after. Thus, while the generated distribution $G_{YX}(Y)$ will look like $X$ at the distribution level, the relationship between a $y_i, y_j$ pair might not match the relationship between the corresponding $G_{YX}(y_i), G_{YX}(y_j)$ pair.

We address this by introducing a loss to explicitly preserve manifold geometry using the same manifold $M$ from the previous section. The manifold geometry loss $L_{MG}$ is thus:
{\small
\begin{align}
    L_{MG} = \sum_{i,j, i \ne j} &(D(M_{x_i}, M_{x_j}) - D(M_{G_{XY}(x_i)}, M_{G_{XY}(x_j)}))^2 \nonumber
\end{align}
}
where $M_{x_i}$ is the representation of $x_i$ on the manifold $M$ and $D$ is a distance function, here chosen to be Euclidean distance on the manifold. We use a coefficient $\lambda_{MG}$ to control the emphasis placed on this term in relation to the importance sampling GAN loss, which we choose to be $.1$ everywhere. 

\section{Experiments}

\begin{figure}
    \centering
    \includegraphics[width=.8\linewidth]{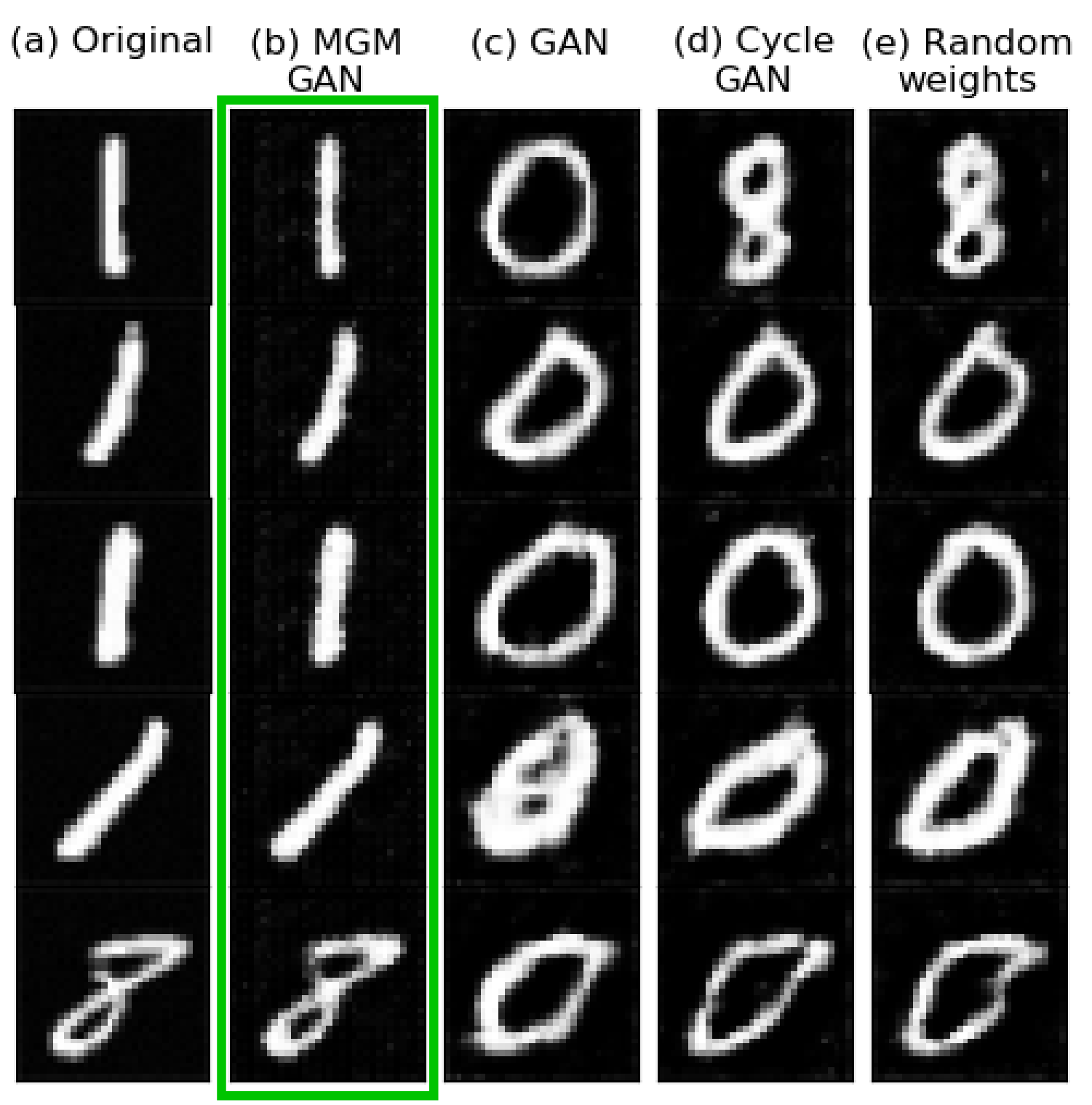}
    \caption{MNIST data where each domain had all digits included, but one domain had ones oversampled and the other domain had zeros oversampled. The GAN wants to turn original digits that are not zeros (first column) into zeros, to match the target density. Since the datasets are just different samples from the same manifold, only MGM GAN correctly aligns them with the identity function.}
    \label{fig:mnist}
\end{figure}

\begin{figure*}
    \centering
    \includegraphics[width=\textwidth]{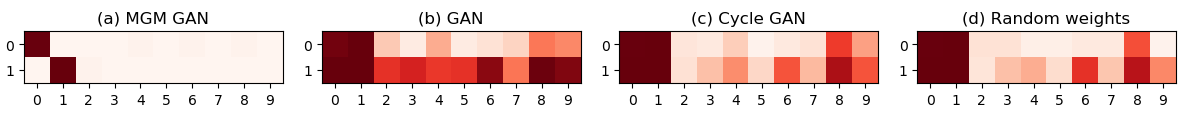}
    \caption{MNIST confusion matrix showing which digits get mapped to by first domain zeros (top row) and second domain ones (bottom row). All but the MGM GAN have to redistribute the oversampled density to other digits, preventing a one-to-one mapping of zeros-to-zeros and ones-to-ones. }
    \label{fig:mnistconfusion}
\end{figure*}

In this section, we start by demonstrating an example of generating from the geometry. Then, we experiment on mapping between domains on: simulated Gaussian mixture models, sampling from the canonical MNIST images, and mass cytometry on T cell development in the mouse thymus, which measure the abundance of various proteins in individual cells. We compare the performance of the MGM GAN to both a traditional \textit{GAN} and a \textit{cycle GAN}. To illustrate the importance of our specific technique for calculating the weights in importance sampling, we then also compare to our model, except using weights that are randomly generated from a uniform distribution (\textit{random weights}) instead. Further implementation details are in the supplemental.

\subsection{Simulated data}
\subsubsection{Geometry Generation}
We first consider an experiment generating from the manifold geometry. The data in this experiment was simulated from a two-dimensional Gaussian mixture model consisting of three Gaussians sampled at different frequencies with a small number of points transitioning between them (Figure~\ref{fig:artificial2}a). The traditional GAN penalty indeed teaches the generator to sample from the density, as can be seen in the kernel density plot in Figure~\ref{fig:artificial2}b, which is dominated by the largest population and misses the transition points. The MGM GAN's importance sampling upweights these points in the low-density region and downweights the points in the high-density region, allowing it to generate evenly over the geometry of the data (Figure~\ref{fig:artificial2}c).

\begin{figure*}
    \centering
    \includegraphics[width=\textwidth]{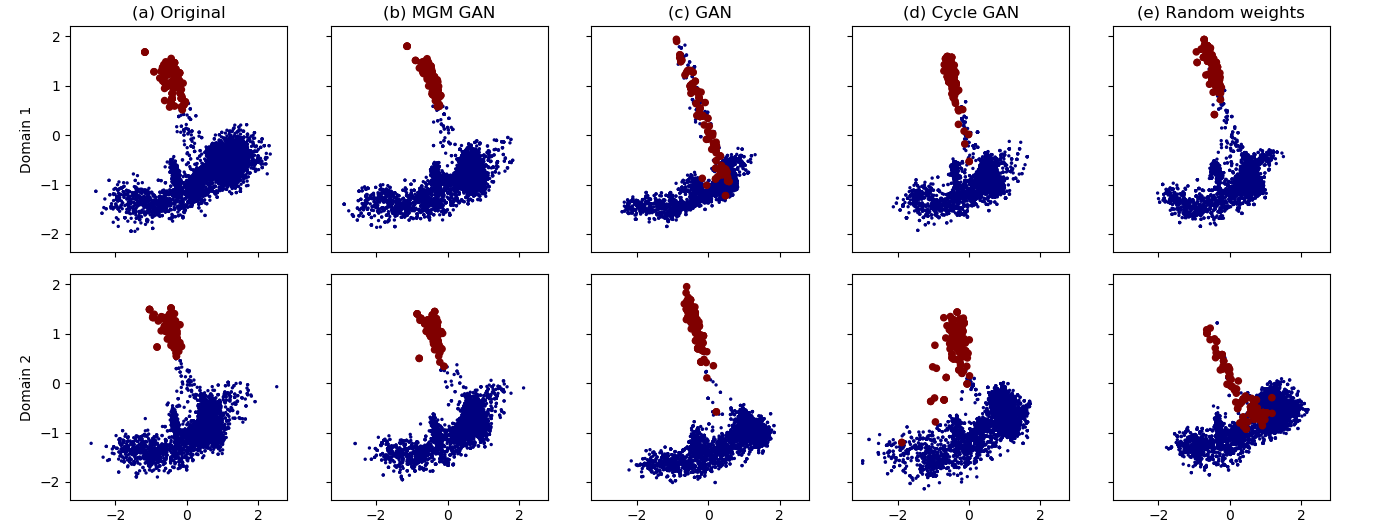}
    \caption{Biological data from \cite{setty2016wishbone}, where there are differences in the frequencies of some cell types between the batches we wish to align. The population of CD25+ cells is highlighted in red. Without the importance sampling and manifold geometry terms of the MGM GAN, some of these cells in the source batch get mapped to other cell types in the target batch. This is due to their different sampling rate between the batches, despite there being no batch effect on them. The MGM GAN preserves the geometry of the original manifold as much as possible while still aligning it to the target manifold.}
    \label{fig:setty}
\end{figure*}
\begin{figure}
\centering
\begin{tabular}{|c|c|c|}
\hline
F-score             & Domain 1 & Domain 2 \\ 
\hline
\hline
MGM GAN              &     0.98 &     0.99 \\ 
GAN                  &     0.27 &     0.47 \\ 
Cycle GAN            &     0.44 &     0.46 \\ 
Random weights       &     0.45 &     0.50 \\ 
\hline
\end{tabular}
\caption{The F-scores for all of the models on MNIST show that the MGM GAN is the only model that can preserve similarities between datasets while being robust to changes in density along the manifold. In the other models, their need to transform one dataset distribution into another data distribution forces them to also change the class of some images, despite them being present in both datasets.}
\label{tab:mnistfscore}
\end{figure}

\subsubsection{Unsupervised Domain Geometry Mapping}
Next, we create two domains out of mixtures of three different Gaussians, but with one of the Gaussians having a minor shift needing alignment. This unsupervised domain mapping presents a significant challenge for traditional GANs, though, because the two domains sample each Gaussian at different frequencies: $0.07$, $0.21$, and $0.72$ respectively for the first domain, and $0.72$, $0.07$, and $0.21$ respectively for the second domain (Figure~\ref{fig:artificial}a).

The traditional GAN penalty prevents aligning the two domains such that the shared Gaussians are aligned together, since their densities along the manifold are different (Figure~\ref{fig:artificial}c-e). In contrast, the importance sampling weighting in the MGM GAN balances the densities, allowing the generator to converge to this alignment (Figure~\ref{fig:artificial}b). Furthermore, without the manifold geometry loss, points that are originally not part of the same Gaussian are mapped to the same Gaussian. With this loss, the MGM GAN preserves the relationships between points before and after mapping, keeping the two representations (one in each domain) of similar points similar and different points different.

\subsection{MNIST}
We next consider an experiment on image data from the canonical MNIST dataset~\cite{lecun1998mnist}. We form the first domain by taking a random sample of $1000$ of each digit except for the digit zero, of which there are $10000$. For the second domain, we do the same with the digit one oversampled. Thus, even though the manifold for the two domains cover the same support, the density along the manifold is different in each domain. In fact, the ones in the first domain are exact elements in the second domain, so it would be desirable to align the two domains such that the class of the elements does not change.

As expected, the traditional GAN loss prevents these models from finding an alignment that preserves the digit identity across domains. Since the ones are oversampled in domain two, most of them get turned into other digits in the other domain (Figure~\ref{fig:mnist}c-e). The oversampling of the zero in the other domain also forces the GANs to create zeros out of other digits, to recreate their abundance in the target domain (last row in Figure~\ref{fig:mnist}c-e).

The MGM GAN importance sampling compensates for the differing densities, and allows the identity function to be a possible local optima for the GAN loss. The manifold geometry further encourages similar original images to be similar after mapping, and different images to be different after mapping. As Figure~\ref{fig:mnist}b shows, this allows the MGM GAN to preserve the identity of the digit through domain transfer.

To quantitatively assess the performance, we consider a slightly different version of the classical task of domain adaptation, which we term \textit{unsupervised domain adaptation}. In traditional domain adaptation, we have labels in one domain and wish to map points to another domain where we have no or few labels. The goal is to classify points in the target domain accurately. Unsupervised domain adaptation is harder because we do not presume to have labels in either domain. Instead, we wish to use unsupervised learning to align the data such that the class of a point is preserved by the mapping.

To evaluate performance at unsupervised domain adaptation, we use the ground truth labels in each original domain and a nearest neighbor classifier to assign labels to generated points in the target domain. We emphasize that these labels are used to score the models, but are not available to them during training. The number of each oversampled digit (zero for the first domain and one for the second domain) that gets mapped to each other digit is shown in Figure~\ref{fig:mnistconfusion}. In the top row, we see the other GANs have to change many of the oversampled zeros to other digits, since the second domain has fewer zeros. This is notable because these zeros are \textit{elements in the other domain}, but are getting changed in the domain mapping anyway. The same happens for the oversampled ones in the other domain. The MGM GAN balances these different densities and consequently performs significantly better at our unsupervised domain adaptation task.

To measure the performance quantitatively, we use F-scores, which deal with the class imbalance by incorporating both precision and recall within each domain~\cite{van1979information}. Scores are reported in Table~\ref{tab:mnistfscore}, where we see as expected, the MGM significantly outperforms the traditional GANs in both domains with scores of $0.98$ and $0.99$, respectively.

\subsection{Biological Data}
In this section, we highlight the importance of modeling the manifold geometry rather than the data density on a real dataset of biological measurements. The data consists of measurements of T cell development in mouse thymus from two individuals, downloaded from \cite{setty2016wishbone}. We would like to integrate measurements from both individuals together, so that further analysis can evaluate both of them together. However, there are multiple sources of variation between the samples that preclude naively combining them.

There are two categories of variation to consider: the first of which we want to correct (instrument error) and the second of which we want to be robust to (differential sampling).
The first, instrument error, is inevitable when running complex machinery as in mass cytometry~\cite{shaham2017removal}. Calibration, amount of reagent, and environmental conditions all can have an effect on measurements, so whenever two samples are compared, these differences need to be reconciled. Most often, this can seen in the existence of a part of the space with points from one sample but not the other~\cite{johnson2007adjusting}. A desirable alignment of the two datasets would correct these differences in support.

The second, differential sampling, though, should not be corrected. While shifts in the support between samples are more likely to be instrument error, we fully expect cell types to be present at different frequencies in different samples. This expectation motivates our need to align manifolds without matching density along the manifolds. For example, we want Cell Type A in one domain to align to Cell Type A in the other domain. The traditional GAN loss would prevent this if Cell Type A is more abundant than Cell Type B in one domain, but the opposite is true in the other domain.

\begin{figure}
    \centering
    \includegraphics[width=.9\linewidth]{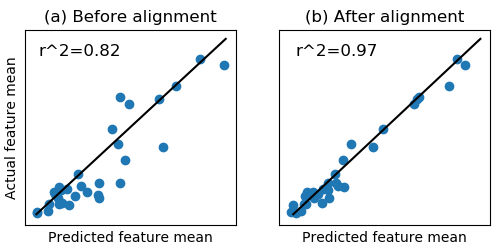}
    \caption{The MGM GAN aligned the samples such that the significant instrument error in measured CD3-CD8+GATA3+ cells is corrected. The mean abundance of each protein for this cell type was very different in the two samples, but after alignment these values correspond significantly better.}
    \label{fig:settycorr}
\end{figure}

The two samples we wish to align consist of $N_1=250170$ and $N_2=220076$ cells, respectively, both having measurements of the abundance of $D=31$ different proteins. An embedding of each sample can be seen in Figure~\ref{fig:setty}a. A difference in geometry between the samples can be seen by examining a particular cell type that is important in the study of mouse development~\cite{vosshenrich2006thymic}. As a part of normal thymus development in mice, cells that are low in a protein called CD3 and high in a protein called CD8 express a protein called GATA3 (CD3-CD8+GATA3+ cells)~\cite{tai2013gata}. In the second sample, these cells make up $34\%$ of all measured cells. In the first sample, they make up just $0.07\%$ of all cells. Moreover, their abundance of other proteins are completely different in the first sample. For example, $80\%$ of these cells are also low in a protein called BCL11b in the second sample, while there are none of the cells at all in the first sample. We would first expect the alignment of the two manifolds to better match this cell type across the samples.

In Figure~\ref{fig:settycorr}, we compare the mean abundance of each protein for CD3-CD8+GATA3+ cells in the original first sample and the second sample. Then, we make this comparison between the transformed first sample and the original second sample. There we see that the transformation significantly improves the accuracy of the alignment, as we would desire, increasing the $R^2$ from $0.82$ to $0.97$.

This first population confirms the MGM GAN's ability to make changes to the geometry of the manifold in order to convincingly generate points in the opposite domain. However, these clear differences in the part of the data space covered by each of the two samples was corrected by all of the GANs, due to the presence of the adversary. The further challenge is to ensure that areas of the geometry that have no batch effect, but possibly have different densities, are not unnecessarily changed. An illustration of the importance of the MGM GAN in such cases is evident when looking at what the transformations did to second cell type population. In both samples, there exists a cell type high in the protein CD25 (CD25+ cells)~\cite{mousecdchart} with no difference in their expression between samples. We would like our alignment to preserve these cells after the transformation. However, they are present in different proportions in the two samples. This means the generator cannot learn an alignment with a one-to-one mapping of CD25+ cells between the samples, as the discriminator would be able to classify this part of the space as preferentially belonging to true samples from one domain or generated samples from the other domain.

\begin{figure}
\centering
\begin{tabular}{|c|c|c|}
\hline
F-score             & Domain 1 & Domain 2 \\ 
\hline
\hline
MGM GAN              &     0.96 &     0.91 \\ 
GAN                  &     0.26 &     0.90 \\ 
Cycle GAN            &     0.92 &     0.71 \\ 
Random weights       &     0.76 &     0.23 \\ 
\hline
\end{tabular}
\caption{F-scores for CD25+ cells in the two samples in the mouse thymus data. This cell type appears in different frequencies in the two samples, so only the MGM GAN's importance sampling optimizes the discriminator with a mapping that preserves the labels across the domains. The other GANs move cells around to match density along the manifold, aligning these CD25+ cells with other completely different cell types.}
\label{tab:settyfscore}
\end{figure}

The MGM GAN's importance sampling balances the differential frequencies and allows a mapping that preserves the CD25+ cells to optimally fool the discriminator. Table~\ref{tab:settyfscore} shows F-scores for CD25+ cells, and we can see the traditional GAN loss forces the other models to move cells around to match the densities along the manifold. As a result, these CD25+ cells are aligned with different cell types, introducing error into any later analysis that uses the aligned data.

\section{Discussion}
In this work we have introduced a novel GAN framework for generating from the manifold geometry. We demonstrate it both in the context of a single GAN generating a single domain and in the context of unsupervised domain mapping. We contribute a re-casting of the traditional density formulation of domain mapping into one of manifold geometry alignment. We model the geometry with an importance sampling technique that weights points based on their density on the manifolds and a novel manifold geometry loss term. The ability to generate from the geometry of the manifold has widespread usage in biology, where sampling makes the density an unreliable represation of the data.

\bibliography{biblio}
\bibliographystyle{icml2019}

\newpage
\clearpage
\section*{Supplemental}
For the artificial dataset, the autoenecoder had three encoder layers and three decoder layers, with dimensions of $200, 100, 50, 10, 50, 100, 200$, an activation function of leaky ReLU with leak of $0.2$ on all layers except the embedding and the output which had linear activation. The generator had the same structure as the autoencoder. The discriminator had three layers with dimension $200, 100, 1$ with leaky ReLU activation on all layers except the last layer, which had a sigmoid.
For the MNIST dataset, convolutional layers were used both in the autoencoder and the generator. The layers had kernel size $3$, stride length $2$, and equal padding. The U-Net architecture~\cite{ronneberger2015u} of skip connections between the encoder and decoder was used with layers of size $32, 64, 128, 64, 32$. The encoder layers had leaky ReLU activation and the decoder layers had ReLU activation. The last layer used the hyperbolic tangent activation.
For the biological dataset, both the same autoencoder structure and generator structure was used as in the artificial dataset except with wider layers. Layer sizes were $800, 400, 200, 100, 200, 400, 800$. The discriminator had layers of dimension $400, 200, 100, 50, 1$.
For all datasets, a learning rate of $0.001$ was used with minibatches of size $200$. The coefficient for the cyle-consistency loss was $1$ and the identity loss was $0.1$.

\end{document}